\documentclass[sn-mathphys]{sn-jnl}
\usepackage{amsmath,mathtools,enumerate, algpseudocode}
\usepackage[numbers]{natbib}

\jyear{2022}%

\newcommand{\beq}{\begin{equation}}
\newcommand{\eeq}{\end{equation}}
\newcommand{\beqa}{\begin{eqnarray}}
\newcommand{\eeqa}{\end{eqnarray}}

\newcommand{\dd}{d}
\newcommand{\sdef}{\stackrel{\mathrm{def}}{=}}
\newcommand{\Rbb}{\mathbb{R}}
\newcommand{\bfQ}{\boldsymbol{Q}}
\newcommand{\bfD}{\boldsymbol{D}}
\newcommand{\bfw}{\boldsymbol{w}}
\newcommand{\calC}{\mathcal{C}}
\newcommand{\calN}{\mathcal{N}}

\newcommand{\bfR}{\boldsymbol{R}}

\newtheorem{theorem}{Theorem}
\theoremstyle{thmstyletwo}%

\theoremstyle{thmstylethree}%

\makeatletter
\def\algbackskip{\hskip-\ALG@thistlm}
\makeatother
\algrenewcommand\algorithmicdo{} 
\algtext*{EndFor}
\algtext*{EndFunction}

\begin{document}

\title[Generalized PaLD]{Generalized partitioned local depth}

\author*[1]{\fnm{Kenneth S.} \sur{Berenhaut}}\email{berenhks@wfu.edu}
\equalcont{These authors contributed equally to this work.}
\author[2]{\fnm{John D.} \sur{Foley}}\email{foley@metsci.com}
\equalcont{These authors contributed equally to this work.}

\author[1,3]{\fnm{Liangdongsheng} \sur{Lyu}}\email{ll675@cam.ac.uk}

\affil*[1]{\orgdiv{Department of Statistical Sciences}, \orgname{Wake Forest University}, \orgaddress{\street{127 Manchester Hall}, \city{Winston-Salem}, \postcode{27109}, \state{NC}, \country{USA}}}

\affil[2]{\orgname{Metron, Inc.}, \orgaddress{\street{1818 Library St., \# 600}, \city{Reston}, \postcode{20190}, \state{VA}, \country{USA}}}

\affil[3]{\orgdiv{Department of Pure Mathematics and Mathematical Statistics}, \orgname{University of Cambridge}, \orgaddress{\street{Wilberforce Road}, \city{Cambridge}, \postcode{CB3 0WA}, \country{UK}}}

\abstract{In this paper we provide a generalization of the concept of cohesion as introduced recently by Berenhaut, Moore and Melvin [{\em Proceedings of the National Academy of Sciences}, {\bf 119} (4) (2022)]. The formulation presented builds on the technique of partitioned local depth by distilling two key probabilistic concepts: {\em local relevance} and {\em support division}. Earlier results are extended within the new context, and examples of applications to revealing communities in data with uncertainty are included. The work
sheds light on the foundations of partitioned local depth, and extends the
original ideas to enable probabilistic consideration of uncertain, variable and potentially conflicting information.}


\maketitle

\section{Introduction}\label{intro}

Uncovering structural communities and clusters within complex data can be of interest across disciplines. In \cite{bmm22}, the authors harness the richness of a social perspective to derive community network structure in the presence of heterogeneity. Therein, a key concept of locality to a pair of data points is provided, leading to informative measures of (local) depth and cohesion. In this paper, we provide a generalization of this approach, by distilling two key probabilistic concepts: {\em local relevance} and {\em support division}. The approach sheds light on the foundations of partitioned local depth, and removes reliance on static distance comparisons, to enable probabilistic consideration of uncertain, variable and potentially conflicting information.

The notion of {\em local (community) depth} introduced in \cite{bmm22} builds on existing approaches to data depth (see for instance \cite{kleindessner2017lens,zuo2000general}). Partitioning the probabilities defining local depth leads to a quantity referred to as {\em cohesion}, which can be understood as a measure of locally perceived closeness. The resulting framework also gives rise to a natural threshold for distinguishing strongly and weakly cohesive pairs and provides an alternative perspective for the concept of near neighbors. Topological features of the data can be considered via networks of pairwise cohesion, and meaningful structure can be identified without additional inputs (e.g., number of clusters or neighborhood size), optimization criteria, iterative procedures, or distributional assumptions. 
For a review of the general method, referred to as {\em partitioned local depth} (PaLD), see Section \ref{form}; for further details see \cite{bmm22}, and the references therein.

It is crucial to note the importance of accounting for varying local density, particularly in applications involving complex evolutionary processes (see, for instance,  \cite{campello2020density,breunig2000lof, domingues2018comparative, everitt1979unresolved}, and examples in \cite{bmm22}). In \cite{bmm22}, relative positioning is considered through distance comparisons within triples of points, which may be of value in non-metric and high-dimensional settings. 

Now, consider a given finite set of interest, $S$. If, for $x,y,z\in S$, we have definitive answers to questions such as ``Is $z$ more similar to $x$ than to $y$?'', then PaLD community analysis can proceed directly \cite{bmm22,bddp22}.  
Still, these may not be the most informative answers to such queries.  For example, answers might instead have inherent variability, e.g.,  80\% of information available suggests that $z$ is more similar to $x$ than to $y$. It may, on the other hand, be the case that there is some true, definitive answer but this answer is subject to inherent uncertainty.

\begin{figure}[ht]
    \centering
    \includegraphics[width=0.8\textwidth]{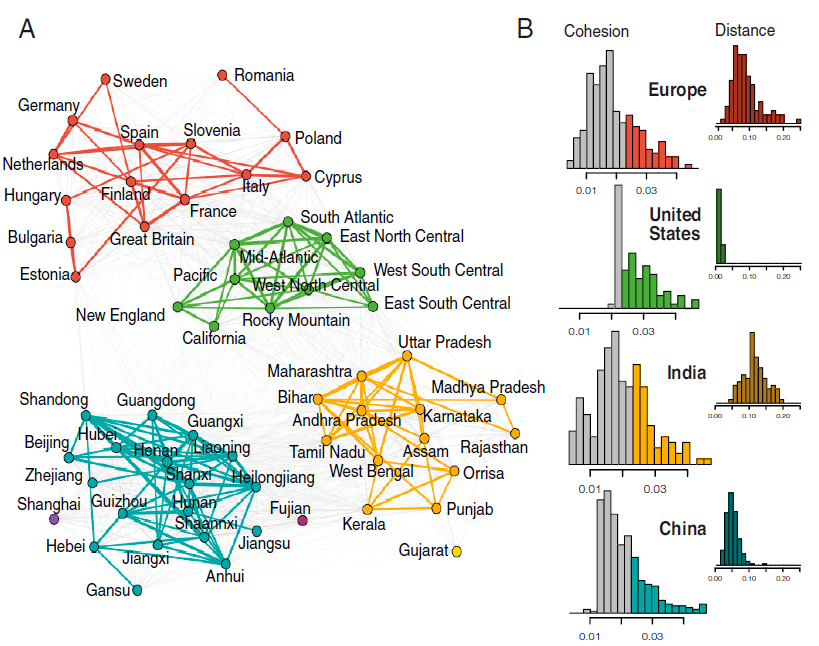}
\caption{Cultural communities from survey data; adapted from \cite{bmm22} with permission. In A, we display the community structure obtained from the cultural fixation index values from [9] for regions within the United States, China, India, and the European Union. In B, we display the distribution of within-group cohesions and distances; colored bars for (mutual) cohesion indicate values above the threshold of 0.0217 (see (\ref{TSd})). Note that distances are brought to comparable levels of cohesion.}
\label{fig_cultural}
\end{figure}

As an example application of PaLD, as introduced in \cite{bmm22}, Fig. \ref{fig_cultural} displays community structure for cultural distance information obtained in \cite{mu20} from two recent waves of the World Values Survey (2005 to 2009 and 2010 to 2014) \cite{in14}. Distances are computed using the cultural fixation index (CFST), which is a measure built on the framework of fixation indices from population biology \cite{bell2009culture,cavalli1994history}. Note that PaLD employs within-triplet comparisons and allows for the employment of such application-dependent, non-Euclidean measures of dissimilarity. 
In Fig. \ref{fig_cultural}A coloured edges correspond to strong mutual cohesion as results from partitioning local depths. For a review of the derivation of such networks see Section \ref{form} below. Histograms for within-group cohesion and distance are provided in Fig. \ref{fig_cultural}B; colored bars for (mutual) cohesion indicate values above the threshold of 0.0217 (see (\ref{TSd}) below). Note that community structure can be identified without additional inputs (e.g., number of clusters or neighborhood size), optimization criteria, iterative procedures, or distributional assumptions.

The data reflect that while, culturally speaking, regions within the United States are far more similar to each other than regions within India, the latter displays similar levels of strong internal cohesion. 

The remainder of the paper proceeds as follows. In Section \ref{form}, we provide some preliminaries and notation, including a review of the development of PaLD as introduced in \cite{bmm22} which highlights its formulation in terms of static dissimilarity comparisons. Section \ref{formu} provides an introduction to the abstracted concepts of local relevance and support division, and the given generalization of PaLD, to incorporate uncertainty, and Section \ref{results} follows with theoretical results on properties of cohesion mirroring those in \cite{bmm22}, for the new scenario. Section \ref{apps} includes mention of potential applications to multiple dissimilarity measures, event-based data and data uncertainty.

We now turn to some preliminaries and notation.

\section{Preliminaries and notation}\label{form}

Suppose $S=\{a_1,a_2,\dots, a_n\}$ is a finite set with a corresponding notion of pairwise dissimilarity or distance $\dd: S \times S \rightarrow \Rbb $.  For any pair $(x,y)\in S\times S$, the set of relevant local data (or {\em local focus}), $U_{x,y}$, is defined to be the set of elements $z\in S$ which are as close to $x$ as $y$ is to $x$, or as close to $y$ as $x$ is to $y$, i.e.  
\beq
U_{x,y} \sdef \{z \in S \mid \dd(z,x) \le \dd(y,x)  ~{\rm or }~ \dd(z,y) \le \dd(x,y)\}.
\label{local_focus}
\eeq
From a social perspective, the set, $U_{x,y}$, local to the pair of individuals $(x,y)$, consists of individuals with alignment-based impetus for involvement in a ``conflict" between $x$ and $y$ (see {\it Social Framework} in \cite{bmm22} for a discussion of the underlying social latent space and related references). In the case of a symmetric distance, $U_{x,y}$ is comprised of those $z$ as close to $x$ or $y$ as they are to each other. The sense of local could be altered depending on applications.

\begin{figure}[H]
    \centering
    \includegraphics[width=0.7\textwidth,trim={2cm 4cm 2cm 5cm},clip]{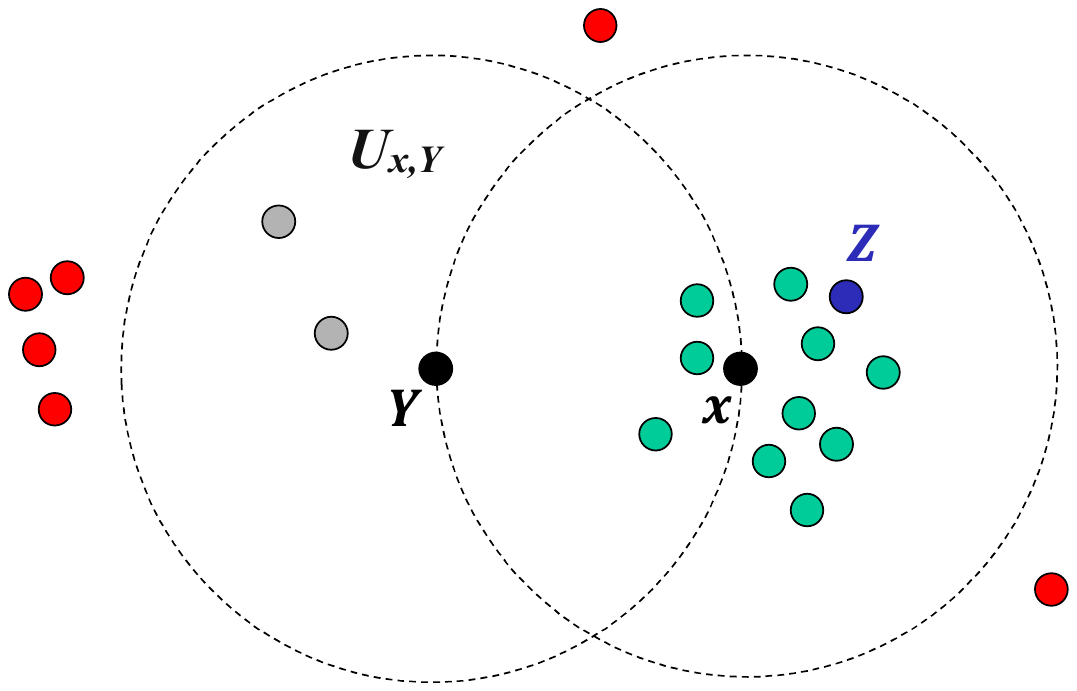}
\caption{The local focus for a fixed point $x$ and a random point $Y$, in two-dimensional Euclidean space. The points in red are outside the focus. Those in green (and $Z$ in blue) are in the focus and closer to $x$, while those in grey are closer to $Y$.}
\label{fig_Uxy}
\end{figure}

The {\em local depth} of $x$, $\ell_S(x)$, is a measure of local support, which leverages the concept of local that is implicit in the definition of  $\{ U_{x,y} \}$:
\beq
\ell(x)\sdef \ell_{S,\dd}(x) = P(\dd(Z,x)<\dd(Z,Y))+ \frac{1}{2} P(\dd(Z,x)=\dd(Z,Y)), \label{local_depth}
\eeq
where $Y$ is selected uniformly at random from the set $S\setminus\{x\}$ and $Z$ is selected uniformly at random from the local set $U_{x,Y}$ (see Fig. \ref{fig_Uxy}). For convenience, the term resolving ties in distance (via coin flip), in (\ref{local_depth}), will be suppressed in what follows.
The important concept of cohesion can then be obtained through partitioning of the probabilities defining $\ell$. In particular, we have that $C_{x,w}$, the {\em cohesion} of $w$ to $x$, is given by
\beq
C_{x,w}\sdef P\left( Z=w,\dd(Z,x)<\dd(Z,Y) \right).  \label{cohesion}
\eeq

The \emph{cohesion network} is the weighted, directed graph with node set $S$ and edge weights $\{C_{x,w}\}$; typically, an undirected version is displayed by considering the minimum of the bi-directional cohesions for each edge pair, with thicker edges depicting larger weights. Unless stated otherwise, we will employ the Fruchterman-Reingold algorithm \cite{fruchterman1991graph} to display cohesion networks. Through cohesion, the dissimilarity measure, $\dd$, is locally adapted, to reflect relative locally-based support (see, for example, Fig. \ref{fig_cultural}). For additional discussion and applications of PaLD, in the context of considerations of data depth, embedding, clustering and near-neighbours see \cite{bmm22}.

As mentioned, though PaLD is formulated in terms of $\dd$, the above definitions in (\ref{local_focus}), (\ref{local_depth}) and (\ref{cohesion}) depend only on \emph{relative} closeness comparisons--e.g., whether $z$ is closer to $x$ than it is to $y$. Thus, as observed in \cite{bmm22,bddp22}, an oracle for triplet comparisons is sufficient to determine the directed cohesion network.  
Note that previous work has suggested that one can often more reliably provide distance comparisons than exact numerical evaluations \cite{kleindessner2017lens, ukkonen2017crowdsourced}. 

As we will see in the remainder of the paper, due to its probabilistic formulation, PaLD is quite readily adapted to allow for uncertainty in dissimilarities.

\section{Generalized PaLD} \label{formu}

Whereas membership of a given $z$ in the local focus $U_{x,y}$ is assumed to be captured by an indicator in $\{0, 1\}$ in (\ref{local_focus}), we will generalize the notion of ``locality'' to a pair $(x,y)$, probabilistically. In a similar manner, support from $Z$ can be formulated stochastically, to give generalized concepts of local depth as in (\ref{local_depth}) and cohesion as in (\ref{cohesion}).          

\vspace{.2in}

\begin{figure}[H]
    \centering
    \includegraphics[width=0.4\textwidth]{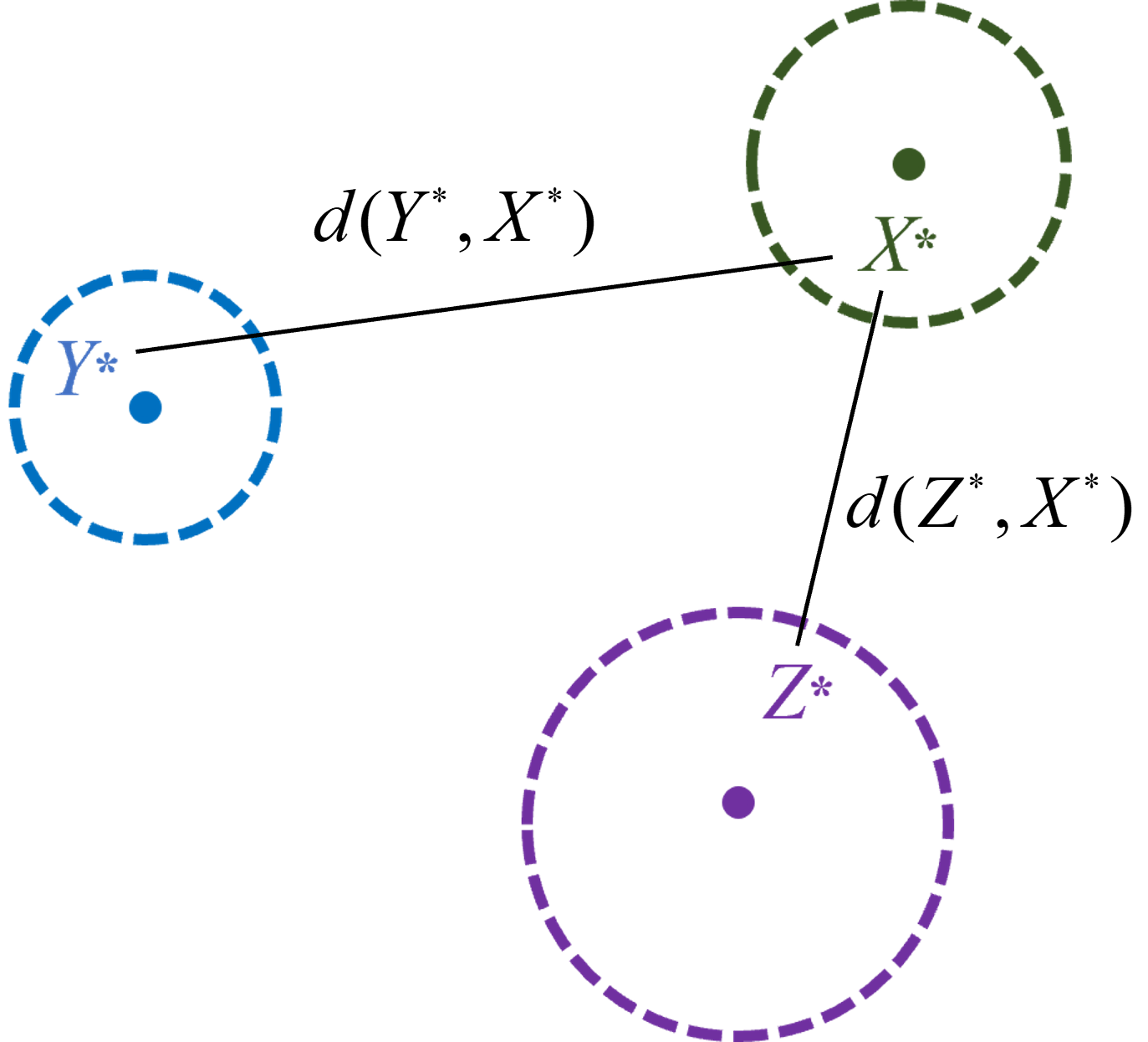}
\caption{Conceptual generative process for \emph{random} triplet comparisons.}
\label{fig_generative}
\end{figure}

\noindent {\bf Example 1.} Before proceeding with formal definitions, for context, consider the simplistic generative process for triplet comparisons depicted in Fig. \ref{fig_generative}. Here we assume that $x$, $y$ and $z$ are fixed, but distance comparisons are based on observed $X^\ast$, $Y^\ast$ and $Z^\ast$ random in neighborhoods about $x$, $y$ and $z$, respectively. We could be interested in uncertain events such as $\dd(Z^\ast,X^\ast)<\dd(Y^\ast,X^\ast)$, say (see Fig. \ref{fig_generative}). Note that static comparisons such as $\dd(z,x)<\dd(y,x)$ may not be fully informative, here. 

\vspace{.2in}

We now introduce the abstracted concepts of local relevance and support division. 

\subsection{Local relevance and support division} \label{lrsd} 
We are interested in generalized definitions of local focus, local depth and cohesion, which reflect uncertainty in dissimilarities. 

For fixed $x,y\in S$, membership in the local focus $U_{x,y}$ can be generalized as follows. 
For each $x,y,z\in S$, define the {\em local relevance of $z$ to the pair $(x,y)$}, $R_{x,y,z}$, as the probability that $z$ is local to the pair $(x,y)$, or more formally
\beq
R_{x,y,z}\sdef P( z \in N (x,y) ),
\label{local_relevance}
\eeq

\noindent where $\calN\sdef \{N(x,y):(x,y)\in S\times S\}$ is a random (pairwise) neighbourhood structure on the set of pairs $S\times S$. Note that we consider the elements in $S$ here as fixed, with no required underlying sense of {\em distance} or {\em position}; stochasticity is provided through $\calN$ (akin to neighborhoods in random graphs). When convenient, we may also consider the full $n \times n \times n$ array of probabilities, $\bfR \sdef [R_{x,y,z}]$. 

To obtain constructions for local depth and cohesion, here, we require a mechanism to sample an element, $Z\in S$, local to $(x,y)$. For this, we consider the process of selecting uniformly at random an element $\tilde{Z}\in S$, and with acceptance probability $R_{x,y,\tilde{Z}}$ taking this as the value of $Z$, repeating the process until a $Z$ is accepted. It is not difficult to see that, for $z\in S$,
\beq
P\left( Z=z \right) = P_{x,y}\left( Z=z \right) = \frac{R_{x,y,z}}{\sum_{w\in S} R_{x,y,w}}.\label{sample_Z}
\eeq
\noindent For fixed $x,y,z\in S$, we may also consider the probability
\beq
Q_{x,y,z}\sdef P( \calC_z(\{x,y\})=x),
\label{support_division}
\eeq
\noindent where  $\calC_z$ is a random choice function, defined on the set of two-element subsets of $S$, i.e., $\calC_z(A)\in A$. Here, $Q_{x,y,z}$ reflects the {\em support division} for $z$ with respect to the pair $(x,y)$. Note that the choice mechanism defined through $\{\calC_z:z\in S\}$ need not be related, per se, to the neighborhood system $\calN$, emphasizing further that the points in $S$ are not required to have fixed position in some underlying, say Euclidean, space. For convenience, we set $\bfQ\sdef[Q_{x,y,z}]$. For general discussion of random choice see for instance \cite{rs06}.

The {\em local depth} of $x$ can then be given by
\beq
\ell(x)\sdef \ell_{S,\bfR,\bfQ}(x) := P(\mbox{$\calC_Z(\{x,Y\})=x$}), 
\label{local_depth_gen}
\eeq  
where, $Y$ is selected uniformly from $S\setminus \{x\}$, and $Z$ is selected with relative weight as in (\ref{sample_Z}).  Likewise, the {\em cohesion} of $w$ to $x$, $C_{x,w}$, generalizes directly as in (\ref{cohesion}):
\beq
C_{x,w}\sdef P( Z=w,\calC_Z(\{x,Y\})=x). \label{Cdefnnew}
\eeq

\noindent Note that the quantity $C_{x,w}$ can be defined independently of $\ell(x)$. We include (\ref{local_depth_gen}), as the work here also generalizes the concept of local depth as defined in \cite{bmm22} (see (\ref{local_depth})).

For examples of computing the arrays $\bfR$ and $\bfQ$ see Section \ref{apps}.

We will assume, throughout, the following basic structural properties on the arrays $\bfR$ and $\bfQ$. Suppose $x,y,z\in S$, 

\vspace{.2in}

\noindent (a) $0 \leq R_{x,y,z},Q_{x,y,z} \leq 1$,    
~~~~~~ (b) $R_{x,y,z}=R_{y,x,z}$,\\
(c)  $Q_{x,y,z}=1-Q_{y,x,z}$,
 ~~~~~~~~~   (d) $R_{x,y,x}=R_{x,y,y}=1$.

\vspace{.2in}

\noindent In (a), we are expressing the fact that the entries in $\bfR$ and $\bfQ$ represent probabilities; in (b), we have that local relevance does not depend on the ordering of $x$ and $y$, (c) reflects the fact that $Z$ supports either $x$ or $y$ (and there is no loss in probability) and (d) states that any individual is locally relevant to any pair in which it is an entry. 

An algorithmic formalization of PaLD, generalized for uncertainty, then follows as in Algorithm \ref{pald:cap} below. The implementation takes the specification of local relevance and support division (through $\bfR$ and $\bfQ$, respectively) as input, to output cohesion. Local depths can be obtained from the row sums of the output matrix, $C$.
\begin{algorithm}
\caption{Generalized partitioned local depth}\label{pald:cap}
 {\bf Input:} arrays $R$ and $Q$ of size $n \times n \times n $:

 \vspace{-0.3in}
\beqa
R_{ijz} &=& P\left( z \in N(i,j) \right) {\rm ~~ and}  \nonumber \\ 
Q_{ijz} &=& P\left( \calC_z(\{i,j\})= i\right)
\nonumber
\eeqa

\begin{algorithmic}
\Function{PaLD}{$R,Q$}
\State{$C \gets {\{0\}}_{i,j=1}^n$}
\For{$i=1$ to $n$} 
\For{$j=1$ to $n$ satisfying $j\neq i$} 
\For{$z=1$ to $n$} 
\State{$C_{iz} \gets C_{iz} + \frac{1}{n-1}
\frac{R_{ijz}}{\sum_{w} R_{ijw}} 
Q_{ijz}$}
\EndFor
\EndFor
\EndFor
\Return{$C$} 
\EndFunction
\end{algorithmic}
 \vspace{0.1in}  {\bf Output:}  matrix of partitioned local depths, $C=[C_{x,w}]$.
 \end{algorithm}

Note that for a given distance function $\dd:S\times S\rightarrow \Rbb$, and $U_{x,y}$ as in (\ref{local_focus}), setting
\beqa
R_{x,y,z} &=& \begin{cases}
1, & \text{if } z\in U_{x,y} \\
0, & \text{otherwise}
\end{cases}, \label{RPaLD}
\eeqa
\noindent and 
\beqa
Q_{x,y,z} &=& 
\begin{cases}
1, & \text{if } \dd(z, x) < \dd(z, y) \\
1/2, & \text{if } \dd(z, x) = \dd(z, y)\\
0, & \text{otherwise}
\end{cases}, \label{QPaLD}
\eeqa

\noindent the 
computation
 of cohesion in \cite{bmm22} is recovered.  
 
Before turning to some applications, we summarize some theoretical results, which generalize and shed light on those given in \cite{bmm22}. 

\section{Results} \label{results}

In this section, we provide results regarding properties of cohesion, mirroring those in \cite{bmm22}, including  (a) {\em dissipation of cohesion under separation}, (b) {\em irrelevance of density under separation}, and (c) {\em dissipation of cohesion for concentrated sets of increasing size}, in the context of uncertainty; proofs can be found in Appendix \ref{proofs}. Throughout, unless stated otherwise, we will assume that the arrays $\bfR$ and $\bfQ$ are fixed, and satisfy the basic assumptions (a)--(d), listed in Section \ref{lrsd}. In addition, $x\in S$ is fixed, $Y$ is selected uniformly at random from $S\setminus \{x\}$ and $Z$ is selected as in Eq. (\ref{sample_Z}).

We begin with three definitions regarding structural properties of the set $S$ with respect to the arrays $\bfR$ and $\bfQ$. The first provides conditions under which two disjoint subsets, $A$ and $B$, of $S$ are {\em sufficiently separated}. In essence, for $c,c^*\in A$ and $d \in B$, $c^*$ is local to the pair $(c,d)$ and fully supports $c$ in that context, while $d$ is not local to the pair $(c,c^*)$.

\vspace{.2in}

\noindent {\bf Definition.} (Sufficiently Separated) Suppose $A,B\subseteq S$. The set $A$ is said to be \textit{sufficiently separated from} $B$ (with respect to $\bfR$ and $\bfQ$) if $A\cap B =\emptyset$, and for all $c,c^*\in A$ and $d\in B$, the following hold:

\noindent ~~~~ (a) $R_{c,d,c^*}=1$,~~~ (b) $R_{c,c^*,d}=0$, 
~~~ (c) $Q_{c,d,c^*}=1$.

\noindent The sets $A$ and $B$ are said to be (mutually) \textit{sufficiently separated} if $A$ is \textit{sufficiently separated from} $B$, and $B$ is sufficiently separated $A$. 

\vspace{.2in}

The second definition is crucial to stating Theorem \ref{irrel_dens}, and addresses equivalence of {\em ordinal} structure for two subsets of $S$ of equal cardinality. 

\vspace{.2in}

\noindent {\bf Definition.} (Equivalence of Ordinal Structure) Suppose two sets $A,B$ satisfy $A=\{a_1,a_2,\dots,a_m\}$, and $B=\{b_1,b_2,\dots,b_m\}$, then $A$ and $B$ are said to have \textit{equivalent ordinal structure}, if they are $(\bfR,\bfQ)$-equivalent, i.e, for $i,j,k\in \{1,2,\dots,m\}$, 
\beq 
R_{a_i,a_j,a_k}=R_{b_i,b_j,b_k} \mbox{    and    } Q_{a_i,a_j,a_k}=Q_{b_i,b_j,b_k}. 
\eeq

\vspace{.2in}

Finally, the following definition suggests a point-like property of one subset, $B \subseteq S$, with respect to another, $A$. In particular if locality to any given pair of elements of $A$ is constant over the set $B$, and all elements of $B$ fully support other elements of $B$ in comparisons with elements of $A$, then $B$ is concentrated with respect to $A$.

\vspace{.2in}

\noindent {\bf Definition.} (Concentrated) Suppose $A,B\subseteq S$, then $B$ is said to be \textit{concentrated with respect to} $A$ (for given $\bfR$ and $\bfQ$), if there exists a function $f: A \times A \rightarrow [0,1]$, such that 
\beq
R_{a,a^*,b} = f(a,a^*)\mbox{~~~~and~~~~}
Q_{a,b,b^*} = 0, 
\eeq

\noindent for $a,a^* \in A$ and $b,b^* \in B$. 

\vspace{.2in}

We have the following results regarding properties of cohesion. Proofs are provided in Appendix A. 

\begin{theorem} \normalsize (Dissipation of cohesion under separation) Suppose $\bfR$ and $\bfQ$ are fixed, $S$ is a disjoint union of $A$ and $B$, and $A$ and $B$ are sufficiently separated with respect to $\bfR$ and $\bfQ$, then the between-set cohesion values are zero, i.e., $C_{a,b}$ = $C_{b,a}=0$ for $a \in A$ and $b \in B$.
\end{theorem}

\begin{theorem} \label{irrel_dens} (Irrelevance of density under separation) Suppose $A=\{a_1,a_2,\dots,a_m\}$ and $A'=\{a'_1,a'_2,\dots,a'_m\}$ have equivalent ordinal structure and $S=A\cup B$ (resp. $S'=A'\cup B$), for some set $B$, where $A$ and $B$ (resp $A'$ and $B$) are sufficiently separated. Then for any $1 \leq i,j \leq m$, $C_{a_i,a_j} = C_{a'_i,a'_j}$, i.e., the corresponding (within-set) pairwise cohesion values are equal.  
\end{theorem}

\begin{theorem} (Dissipation of cohesion for concentrated sets of increasing size) Suppose $S$ is a disjoint union of $A$ and $B$, and $B$ is sufficiently separated from, and concentrated with respect to $A$. Then, for $a \in A$ and $b \in B$, the cohesion of $b$ to $a$ tends to zero as $\lvert B\rvert$ tends to infinity.
\end{theorem}

\vspace{.2in}

The next result follows from the probabilistic definition of local depth along with the assumptions (c) and (d), from Section \ref{lrsd}, namely 
\beq
Q_{x,y,z}=1-Q_{y,x,z} \mbox{~~~~and~~~~} R_{x,y,x}=R_{x,y,y}=1. \nonumber
\eeq

\noindent Here, the first assumption provides conservation of probability and the second guarantees proper selection of $Z$.

\begin{theorem} (Conservation of Cohesion) We have 
\beq
\frac{n}{2}=\sum_{x\in S} \ell_{S,\bfR,\bfQ}(x) =\sum_{x,w\in S}C_{x,w}.
\eeq
\end{theorem}

Finally, in \cite{bmm22}, a threshold distinguishing strong from weak cohesion is provided. In particular, define 

\beq
T_{S,\dd}\sdef P(Z=W,\dd(Z,X)< \dd(Z,Y))=\frac{1}{2n}\sum_{x\in S}C_{x,x},
\label{TSd}
\eeq

\noindent where $X, Y, Z$ and $W$ are selected uniformly at random from $S$, $S\setminus \{x\}$, $U_{X,Y}$ and $U_{X,Y}$, respectively. For the generalization provided here, the analogue of the final equality in (\ref{TSd}) no longer necessarily holds, but we do have the following. 

\begin{theorem} \label{thm5}
Set $T\sdef T_{S,\bfR,\bfQ}=P(Z=W, \calC_Z(\{X,Y\})=X)$. Then 
\beq
T\leq \frac{1}{2n}\sum_{x\in S}C_{x,x}. \label{thresholdbd}
\eeq
\end{theorem}

\vspace{.2in}

\noindent Key to the proof of Theorem \ref{thm5} is the fact that here, in place of $P(Z=W)= P(Z=X)$ as available in \cite{bmm22}, we only have $P(Z=W)\leq P(Z=X)$ due to selection being dependent on the local relevance array, $\bfR$. One would have equality in the case of a $(0,1)-\bfR$ array, even in the presence of flexibility allowed for support division. 

We now turn to discussion of some potential applications. 

\section{Applications}\label{apps}

In this section we consider applications of the concepts of local relevance and support division in revealing community structure in complex data. Results follow upon determination of the arrays $\bfR$ and $\bfQ$. Importantly, the foundational framework from \cite{bmm22} carries over (see the results in Section \ref{results}). Note at the outset that the perspective on community structure
developed in [1] (and extended here) is quite distinct from clustering. See {\em Discussion and Conclusions} in \cite{bmm22} for further details on implications of the underlying PaLD perspective in this context.

\subsection{Combining multiple dissimilarity measures} \label{cultural}

Our ability to reason directly from local relevance and support division allows for flexibility to combine multiple, possibly conflicting, dissimilarity measures. Instead of linearly combining such measures to form one, say, we can proceed probabilistically.

\vspace{.2in} 

\noindent {\bf Example 2.} Recall the cultural values data considered earlier in Fig. \ref{fig_cultural}. Distances for politically-related questions are provided at \cite{CD23} for the dimensions of {\em Politics}, {\em Democracy}, {\em Egalitarianism}, {\em Conservatism}, {\em Neoliberalism}, {\em Authoritarianism}, {\em Libertarianism}, {\em Change}, and {\em Social}. 
We will focus, here, on the subset of 25 European countries (out of 27) for which there is complete data for these dimensions. Define the respective resulting distance matrices as $\bfD_1,\bfD_2,\dots,\bfD_9$.

Consider two potential methods for combining the pairwise distance information for the countries, to obtain cohesion networks. In one, we could obtain a single distance matrix, via simple linear weighting, i.e., for a non-negative weight vector, $\bfw=(w_1,w_2,\dots,w_9)$, satisfying $\sum_i w_i=1$
\beq
\bfD^*_{\bfw}\sdef w_1\bfD_1+w_2\bfD_2+\cdots+w_9\bfD_9.\label{Dstar}
\eeq
\noindent and proceed with PaLD, as in \cite{bmm22}.
Alternatively, we could obtain respective arrays $\bfR_1,\bfR_2,\dots,\bfR_9$ and $\bfQ_1,\bfQ_2,\dots,\bfQ_9$, (as in (\ref{RPaLD}) and (\ref{QPaLD})), and weight these to give
\beq
\bfR^*_{\bfw}\sdef w_1\bfR_1+w_2\bfR_2+\cdots+w_9\bfR_9,\label{Rstar}
\eeq
\noindent and
\beq
\bfQ^*_{\bfw}\sdef w_1\bfQ_1+w_2\bfQ_2+\cdots+w_9\bfQ_9.\label{Qstar}
\eeq

Note that the $(i,j,k)$-entry in the array $\bfR^*_{\bfw}$ can be viewed as 
\beq
(\bfR_{\bfw}^*)_{i,j,k}=w_1(\bfR_1)_{i,j,k}+w_2(\bfR_2)_{i,j,k}+\cdots+w_9(\bfR_9)_{i,j,k},\label{Rstarijk}
\eeq
expressing the fact that $(\bfR_{\bfw}^*)_{i,j,k}$, the $(i,j,k)$-entry in the array $\bfR^*_{\bfw}$, is the probability that when a dimension, $\delta$, is selected according to the distribution $\bfw$, we have $a_k \in N(a_i,a_j)$ under that respective distance. Since relations are considered solely under individual distance matrices, this allows dimensions of differing scales and data types to be readily considered.

\begin{figure}[H]
    \centering
  \includegraphics[width=0.8\textwidth,trim={0cm 12cm 0cm 0cm},clip]{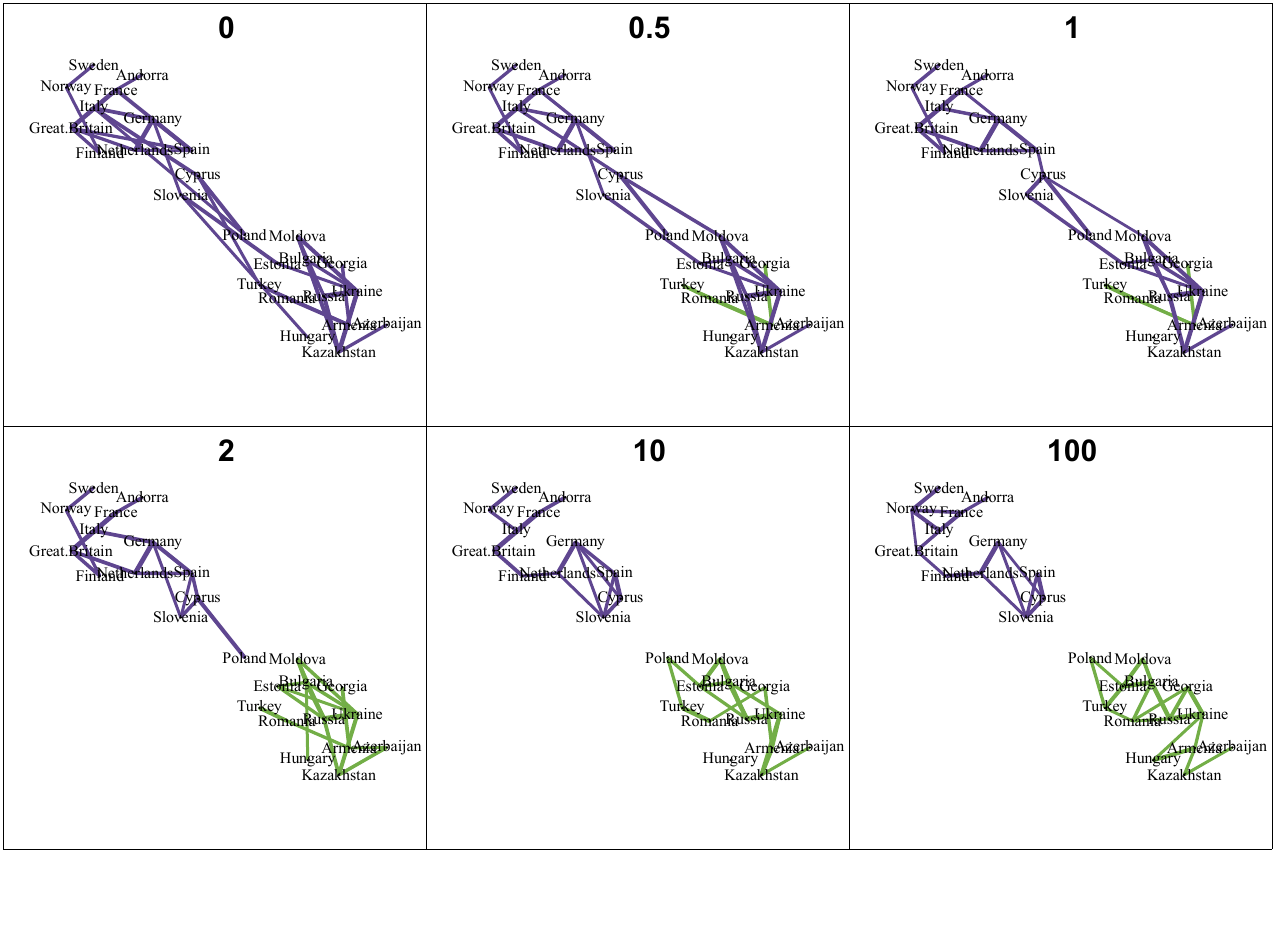}
\caption{Cohesion networks based on $D^*$ as the relative weight on the {\em Social} dimension increases through the values $0,0.5,1.0,2.0,10,100$. Ties above the threshold in (\ref{TSd}) are displayed. The layout for each plot is that based on the {\em Social} dimension in isolation.}
\label{fig_one_dist}
\end{figure}

For illustration, Fig. \ref{fig_one_dist} contains a display of the cohesion networks resulting from weight vectors where the relative weight on $\bfD_9$ (for the {\em Social} dimension, with equal weights for others), increases through the values $0,0.5,1.0,2.0,10,100$. 

\begin{figure}[H]
    \centering
    \includegraphics[width=0.8\textwidth,trim={0cm 12cm 0cm 0cm},clip]{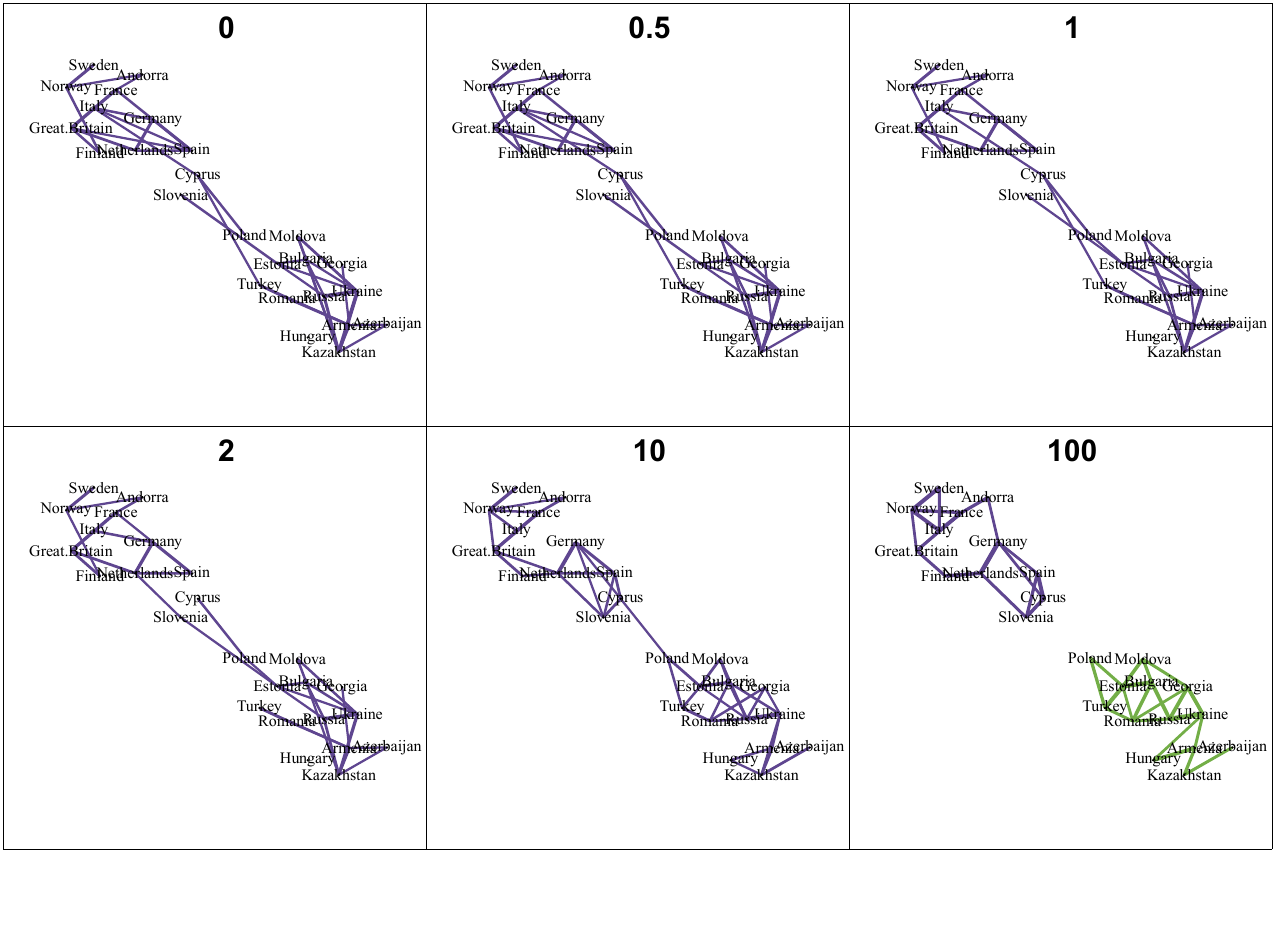}
\caption{Cohesion networks based on $R^*$ and $Q^*$ as the relative weight on the {\em Social} dimension increases through the values $0,0.5,1.0,2.0,10,100$. Ties above the threshold in (\ref{thresholdbd}) are displayed. The layout for each plot is that based on the {\em Social} dimension in isolation.}
\label{fig_combine_R_Q}
\end{figure}

Fig. \ref{fig_combine_R_Q} contains a display of the cohesion networks resulting from weight vectors where the relative weights on $\bfR_9$ and $\bfQ_9$ increase through the same values $0,0.5,1.0,2.0,10,100$. Note some potential added stability in the cohesion network, as the weight on the {\em Social} dimension increases.

Typically one may choose to use uninformative uniform weights $\{w_i\}$, in obtaining $\bfR^*$  and $\bfQ^*$. Further considerations of combining measures, and weight selections is work in progress. For discussion of combining dissimilarity measures from mixed-type data in the context of clustering see for instance \cite{cpm22}, and the references therein. Note as mentioned the probabilistic framework here maintains the properties of PaLD, and avoids need to consider standardization choices within dimensions. 

\subsection{Event-based data} \label{event}

Another potential application of the concepts of local relevance and support division is to similarity determined by multiple events. For instance, consider a set $S$ of individuals, where for each pair $(x,y)\in S\times S$, we have a set of dissimilarities $A_{x,y}$, each with non-zero cardinality $n_{x,y}\sdef\lvert A_{x,y}\rvert$. Note that it is not necessary that $n_{x,y}$ be constant over pairs $(x,y)$. There are several ways in which such similarities might arise. Consider the following example. 

\vspace{,2in}

\begin{figure}[H]
    \centering
    \includegraphics[width=0.5\textwidth,trim={2cm 2cm 2cm 2cm},clip]{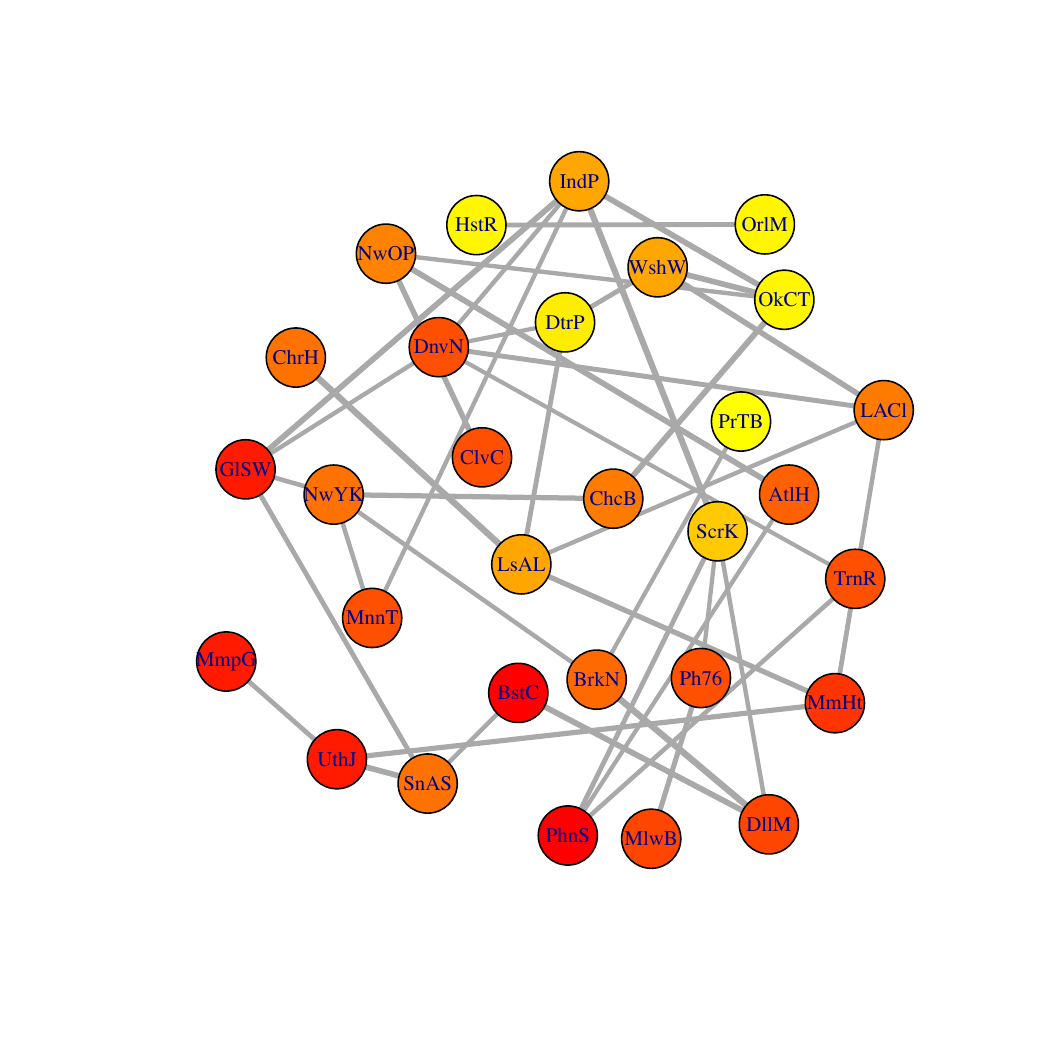}
\caption{The cohesion network for the 2021--2022 NBA basketball season based on proportional point differentials. Shading of nodes is according to mean proportional point differential; the highest is for the Phoenix Suns (0.034122; red) and lowest is for the Portland Trail Blazers (-0.04019; yellow). Edge-width is proportional to mutual cohesion. Note that team names have been abbreviated for display.}
\label{fig_NBA}
\end{figure}

\noindent {\bf Example 3.} Suppose we have competing entities for which multiple events determine pairwise distance, e.g. firms in different markets or competitors in an athletic context. For fixed $x,y,z\in S$, values for $Q_{x,y,z}$ and $R_{x,y,z}$ can be determined as probabilities through random (potentially weighted) selections from $A_{x,y}$, $A_{y,z}$ and $A_{x,z}$. For concreteness, in Fig. \ref{fig_NBA}, we consider a cohesion network based on pairwise similarities determined by competitiveness in games played between teams during the 2021--2022 season of the {\em National Basketball Association} (NBA). Here, dissimilarity in a particular event (game) was determined as the proportion of (absolute) point differential to overall game point total. For instance, a score of 110-90 would result in a non-competitiveness score of $\lvert 110-90\rvert/(110+90)=0.10$. Note that, in this case, the values of $\{n_{x,y}\}$ vary between 2 and 4. The edges corresponding to {\em strong} pairwise cohesions above the threshold bound in (\ref{thresholdbd}) are displayed, in  Fig. \ref{fig_NBA}. Note that the figure shows a general gradient from weaker teams at the top right to stronger teams at the bottom left. The largest cohesion is between the Dallas Mavericks and Brooklyn Nets, while the lowest is between the Phoenix Suns and the Charlotte Hornets. Some weaker teams display relative competitiveness with stronger teams head-to-head, such as the Detroit Pistons with the Denver Nuggets (two games with proportional point differentials of 6/(117+111) and 5/(105+110)). 

Further applications could include any instances where event results determine distances. Similar ideas could also be used, when the events are drawn from sampling pairs of entities (and measuring dissimilarities) over time.

The final application included here is a line of potential further work, with applicability in the context of addressing discrete jumps in cohesion, adapting to cases where there is known levels of data precision and considerations of structural persistence.

\subsection{Data uncertainty}\label{uncertainty}

If 
we have information regarding data uncertainty, then, for fixed $x,y,z\in S$, it is possible to adjust $R_{x,y,z}$ and $Q_{x,y,z}$ from indicators, as in (\ref{RPaLD}) and (\ref{QPaLD}), directly to probabilities. That is, $R_{x,y,z}$ could reflect the probability of membership of $z$ in the local focus of $(x,y)$ and $Q_{x,y,z}$, the probability of $z$ being closer to $x$ than to $y$.  More generally, adjustment for various sources of uncertainty becomes possible and has the potential advantage of making cohesion continuous in the data.      

\vspace{.2in}

\noindent {\bf Example 4.}
If we assume a sufficiently simple model, exact calculations of $\bfR$ and $\bfQ$ are relatively straightforward.  
Suppose that $\epsilon>0$ is fixed and each $a\in S\subseteq \mathbb{R}$ has random associated value $A^\ast\in S^\ast\subseteq 	\mathbb{R}$, uniformly distributed in an $\epsilon$-ball centered at $a$. Here $S^\ast$ is the set of associated values. We can then compute the arrays $\bfR$ and $\bfQ$, in terms of corresponding entries in the set $S^\ast$. If $\epsilon$  (under uniformity) accurately reflects measurement uncertainty, then cohesion can be more faithfully modeled. 

In this scenario, cohesion can be seen to be stable with respect to small changes in the data. Rather than having discrete jumps, due to discontinuities in (\ref{RPaLD}) and (\ref{QPaLD}), a positive $\epsilon$ (e.g., reflecting the precision used to store the data) makes cohesion a continuous function of $S$. For instance, for $x,y,z\in S$, with $x<y<z$, 
if $z$ is sufficiently close to $y$ (relative to $\epsilon$), then $z$ is in the local focus with probability one, i.e., $R_{x,y,z}=1$.  However, if $z$ is gradually increased (moving farther from $y$), $R_{x,y,z}$ transitions to the value zero. 

 Finally, we can consider how cohesion varies as $\epsilon$ increases, in a manner similar to persistent homology \cite{wasserman2018topological}.
It is currently work in progress to consider higher dimensional scenarios and more complex settings. For consideration of clustering in the context of uncertain data, see for instance \cite{skezsz15,gpt08}. 

\section{Conclusion}\label{sec13}

The generalization of partitioned local depth, developed here, enhances PaLD's theoretical underpinnings and broadens the potential application of cohesion to complex data for which there may be uncertain, variable or conflicting information. 

Two key probabilistic concepts, local relevance and support division, are introduced leading to an extended probabilistic framework for revealing communities in data.

Base properties of the resulting cohesion values have been proven and initial potential applications in the contexts of multiple dissimilarity measures, event-based data and data uncertainty are discussed. Several questions remain, as suggested throughout the manuscript. We have provided examples of applications in Section \ref{apps}, but general determination of arrays $\bfR$ and $\bfQ$ (and their impact for representative community structure) is important for future work. It is hoped that the present work may lead to further consideration of communities in data. 

\backmatter

\bmhead{Acknowledgments}
The authors thank Katherine Moore, Richard Darling, several individuals at Metron, Inc., and others for stimulating discussions on communities in data.   

\begin{appendices}

\section{Proofs of results} \label{proofs}

\noindent {\bf Proof of Theorem 1.}  Suppose $a \in A$ and $b \in B$, $Y$ is selected uniformly at random, and for $s \in S$,
\beq
P(Z=s) = \frac{R_{a,Y,s}}{\sum_{w}{R_{a,Y,w}}}.
\eeq
\noindent Then partitioning according to the location of $Y$, and employing the definition in (\ref{Cdefnnew}), 
\beq
C_{a,b} = P\left( Z=b,  \calC_Z(\{a,Y\})=a, Y \in A\right) + P\left( Z=b, \calC_Z(\{a,Y\})=a, Y \in B\right). \label{partY}
\eeq

\noindent In the case that $Y \in A$, since $A$ is sufficiently separated from $B$, $R_{a,Y,b}=0$. If $Y\in B$, since $B$ is sufficiently separated from $A$, $Q_{a,Y,b}=0$, and hence $C_{a,b}=0$. Similarly, $C_{b,a}=0$. \hfill \qed

\vspace{.2in}

\noindent {\bf Proof of Theorem 2.} Suppose that $1 \leq i, j \leq m$ are fixed and set $x=a_i$ and $w=a_j$ (resp. $x^{\prime}=a_i^{\prime}$ and $\left.w^{\prime}=a_j^{\prime}\right)$. As in (\ref{partY}),
\beq
C_{a,b} = P\left( Z=b, \calC_Z(\{a,Y\})=a, Y \in A\right) + P\left( Z=b,  \calC_Z(\{a,Y\})=a, Y \in B\right). 
\eeq

In the case $Y \in A$ (resp. $Y' \in A'$), since $A$ and $A'$ are separated from $B$, $R_{x,Y,b}=R_{x',Y',b}=0$ for all $b \in B$, and hence, since $A$ and $A'$ have equivalent ordinal structure,
\beq
P\left( Z=w, \calC_Z(\{x,Y\})=x\mid Y \in A\right) = P\left( Z=w',  \calC_Z(\{x',Y\})=x'\mid Y \in A'\right).
\eeq

\noindent On the other hand, if $Y \in B$, then since $A$ is sufficiently separated from $B$, $R_{x,Y,w}=1$ and $Q_{x,Y,w}=1$, and since $B$ is sufficiently separated from $A$, for $b \in B$, $R_{x,Y,b}=1$ and $Q_{x,Y,b}=0$. Therefore,
\beqa
&~&P\left( Z=w,  \calC_Z(\{x,Y\})=x\mid Y \in B\right) = \frac{1}{n} \nonumber \\
&~&~~~~~~~~~~~~~~
= P\left( Z=w',  \calC_Z(\{x',Y\})=x'\mid Y \in B\right).
\eeqa

\noindent Since $P(Y \in A)=P\left(Y \in A^{\prime}\right)$, the result now follows. \hfill \qed

\vspace{.2in}

\noindent {\bf Proof of Theorem 3.} Suppose $a \in A$ and $b \in B$. Since $B$ is sufficiently separated from $A$, $Q_{a,y,b}=0$ for all $y \in B$, and hence, 
\beq
P\left( Z=b, \calC_Z(\{a,Y\})=a, Y \in B\right) = 0.  
\eeq

\noindent For $y \in A$,  since $B$ is concentrated with respect to $A$,
\beq
P\left( Z=b,  \calC_Z(\{a,Y\}=a\mid Y =y\right) \leq \frac{f(a,y)}{\lvert B\rvert f(a,y)}=\frac{1}{\lvert B\rvert}, 
\eeq
\noindent and hence 
\beqa
C_{a,b} &=& P\left( Z=b,   \calC_Z(\{a,Y\})=a, Y \in B\right) \nonumber \\
&~&~~~~~~+ P\left( Z=b,   \calC_Z(\{a,Y\})=a, Y \in A\right)\leq \frac{1}{\lvert B\rvert}, 
\eeqa
\noindent and the results follows. \hfill \qed

\vspace{.2in}

\noindent {\bf Proof of Theorem 5.} The proof follows as in \cite{bmm22}, except, whereas therein $P(Z=W)=P(Z=X)$, here by the assumption (d) in Section \ref{lrsd}, $R_{X,Y,W}\leq R_{X,Y,X}$, and hence
\beq
P(Z=W)\leq P(Z=X)=\frac{1}{n} \sum_x P(Z=x)=\frac{1}{n} \sum_x P(Z=x,  \calC_Z(\{x,Y\})=x).
\eeq

The result follows by employing the definition of $C_{x, x}$, the assumption (b) in Section \ref{lrsd} and leveraging the symmetry in the selection of $X$ and $Y$. \hfill \qed

\end{appendices}

\vspace{.2in}

\noindent {\bf Conflict of interest statement:} On behalf of all authors, the corresponding author states that there is no conflict of interest.

\bibliography{sn-bibliography}

\end{document}